\documentclass[12pt,noinfoline,a4paper]{imsart}
\RequirePackage{amsthm,amsmath,amsfonts,amssymb}
\RequirePackage[authoryear]{natbib}
\RequirePackage[colorlinks,linkcolor=cyan,citecolor=red,urlcolor=magenta]{hyperref}
\usepackage[table,xcdraw]{xcolor}
\usepackage{arydshln}
\usepackage{color,soul}
\usepackage{accents,float,graphicx,subfig,multirow,dcolumn,booktabs,lscape}
\usepackage[a4paper,margin=2.54cm]{geometry}
\usepackage[ruled,linesnumbered]{algorithm2e}
\usepackage{comment}

\pagecolor[rgb]{0.1,0.1,0.1} 
\color[rgb]{0.85,0.85,0.85} 

\startlocaldefs
\numberwithin{equation}{section}

\theoremstyle{definition}

\providecommand{\definitionname}{Definition}
\numberwithin{definition}{section}

\def\spacingset#1{\renewcommand{\baselinestretch}{#1}\small\normalsize}\spacingset{1}

\newcolumntype{.}{D{.}{.}{-1}}
\definecolor{red}{rgb}{1.0, 0.0, 0.0}

\endlocaldefs

\begin{document}

\begin{frontmatter}

\title{Accuracy and stability of solar variable selection comparison under complicated dependence structures\thanksref{T1}}

\thankstext{T1}{Xu would like to thank Google Australia and NICTA for hardware and programming assistance in package optimization and development. Xu also would like to thank Dr. Peter Exterkate, Uni Sydney and Prof. A. Colin Cameron, UC Davis for their valuable advice.Fisher would like to acknowledge the financial support of the Australian Research Council grant DP0663477.}

\begin{aug}
  \author{\fnms{Ning }\snm{Xu,}\ead[label=e1]{n.xu@sydney.edu.au}}
  \author{\fnms{Timothy C.G. }\snm{Fisher}\ead[label=e2]{tim.fisher@sydney.edu.au}}
  \and
  \author{\fnms{Jian }\snm{Hong}\ead[label=e3]{jian.hong@sydney.edu.au}}

  \address{School of Economics, University of Sydney\\ NSW 2006 Australia\\
  \printead{e1,e2,e3}}
\end{aug}

\begin{abstract}
  In this paper we focus on the empirical variable-selection peformance of subsample-ordered least angle regression (Solar) --- a novel ultrahigh dimensional redesign of lasso --- on the empirical data with complicated dependence structures and, hence, severe multicollinearity and grouping effect issues. Previous researches show that Solar largely alleviates several known high-dimensional issues with least-angle regression and $\mathcal{L}_1$ shrinkage. Also, With the same computation load, solar yields substantiali mprovements over two lasso solvers (least-angle regression for lasso and coordinate-descent) in terms of the sparsity (37-64\% reduction in the average number  of selected variables), stability and accuracy of variable selection. Simulations also demonstrate that solar enhances the robustness of variable selection to different settings of the irrepresentable condition and to variations in the dependence structures assumed in regression analysis. To confirm that the improvements are also available for empirical researches, we choose the prostate cancer data and the Sydney house price data and apply two lasso solvers, elastic net and Solar on them for comparison. The results shows that (i) lasso is affected by the grouping effect and randomly drop variables with high correlations, resulting unreliable and uninterpretable results; (ii) elastic net is more robust to grouping effect; however, it completely lose variable-selection sparsity when the dependence structure of the data is complicated; (iii) solar demonstrates its superior robustness to complicated dependence structures and grouping effect, returning variable-selection results with better stability and sparsity. The code can be found at \url{https://github.com/isaac2math}
\end{abstract}


\end{frontmatter}


\spacingset{1.5}


\section{Introduction}

Variable selection is essential in the modern statistics. It has been successfully applied to high-dimensional predictions, dependence structure estimations and structure learning across many fields, including finance, machine learning, biostatistics and signal-processing. In the field of linear modelling, different researchers have introduced different new algorithms (e.g., lasso and variabel screening) with better theoretical properties and the simulation performances. With ever-expanding data dimensions, those new algorithms are designed to select a sparse set of variables that works well for prediction, interpretation and dependence structure estimation. However, due to the complicated dependence structure and, hence, the multicollinearity issue in the real-world data, these new algorithms may not perform well in real-world applications. Hence, it is necessary and still challenging to perform an accurate and stable variable selection under complicated dependence structures and multicollinearity.

To improve the accuracy and stability of variable-selection, \citet{ning2019solar} propose the \emph{subsample-ordered least-angle regression (solar)}, a ultrahigh dimensional redesign of lasso. Solar relies on the average $\mathcal{L}_0$ solution path computed across subsamples and largely alleviates several known high-dimensional issues with least-angle regression and $\mathcal{L}_1$ shrinkage. Usinge examples based on directed acyclic graphs, we illustrate the advantages of solar in comparison to least-angle regression, forward regression and variable screening. Simulations demonstrate that, with the same computation load, solar yields substantiali mprovements over two lasso solvers (least-angle regression for lasso and coordinate-descent) in terms of the sparsity (37-64\% reduction in the average number  of selected variables), stability and accuracy of variable selection. Simulations also demon-strate that solar enhances the robustness of variable selection to different settings of the irrepresentable condition and to variations in the dependence structures assumed in regression analysis. With all the significant improvements in simulations, however, the improvements of solar has not been confirmed on real-world data with complicated dependence structures and multicollinearity issues. In this paper, we focus on the variable selection performance of solar on two real-world data with heavy multicollinearity and complicated dependence structures, the prostate-cancer data (small $p$) and Sydney house data (moderate $p$). Taking two lasso solvers and elastic net as competitors, we show that solar outperform them by reaching a better balance between sparsity and variable-selection stability.

\subsection{Literature review}

As a major issue of linear modelling for decades, multicollinearity can cause problems on classical techniques of linear modelling from different perspectives. Firstly, since linear modelling can be considered as the error minimization in a linear space, the multicollinearity issue will reduce the magnitude of the minimal eignvalue in the linear space, causing different issues on numerical convergence (e.g., the Cholesky decomposition or the gradient descent) and model estimation. Moreover, a severe multicollinearity will amplify the instability of the parameter estimate across samples. For example, the more severe the multicollinearity issue is, the more dramtically the sample regression coefficients will change across samples, implying that it is improbable to interpret the sample regression coefficients reliably and accurately. Furthermore, the multicollinearity issue also causes problems on statistical tests. A severe multicollinearity issue will unnecessarily overamplify the volume of the standard error of regression coefficients. As a result, the finite-sample performance of all the statistical tests that rely on the sample covariance (e.g., the post-OLS t-test or the covariance test of lasso \citep{lockhart2014significance}, the condtional correlation tests of dependence structure estimation \citep{scutari2014bayesian}) will be weakened \citep{farrar1967multicollinearity}. Last but not least, the multicollinearity may also reduce the algorithmic stability of the model \citep{elisseeff2003leave}, which reduce the generalization ability and the prediction ability of the estimated model.

Multicollinearity also affects the reliability of the variable selection algorithms in linear modelling. For example, the lasso regression \citep{tibshirani96} will be unstable if a group of variables are highly correlated to each other \citep{zou2005regularization, jia2010model}. Lasso will randomly select one from the group and drop the other out of the regression model, which is referred to as \textbf{the grouping effect}. For all linear modelling techniques, the variable selection decision is based on the conditional correlation between a covariate $\mathbf{x}_j$ and the response $Y$ while controlling the other covariate. As a result, the grouping effect may well apply to other variable selection methods like the best subsset method (including AIC, BIC and Mallow's $\mathcal{C}_p$), reducing the stability and accruacy of the variable selection in linear modelling.

The consequence of grouping effect and multicollinearity has gone beyond the field of variable selection in linear modelling. Since (i) it is NP-hard to estimate the dependence struture (also referred to as probablistic graph learning) on data with large $p$ \citep{heckerman95, chickering04}; (ii) the dependence structure estimation algorithms typically work on data with large $n$ and very sparse $p$, variable selection methods in linear modelling (e.g., SCAD \citep{fan2001variable}, ISIS \citep{fan2008sure} and different lasso-type estimators \citep{fan2009network}) are frequently used to filter out the redundant variables before estimating the linear dependence structures in biostatistics and machine learning. However, due to the complicated linear structure and, hence, the grouping effect, lasso or other classical varibale selection methods may randomly drop some of the highly correlated variables, resulting in the omissions of important variables in the linear structure.

Different attempts have been made to reduce the effect of multicollinearity. For a more stable regression coefficients estimate, \citet{hoerlkennard70} apply the Tikhonov regularization to OLS, resulting in the Ridge regression. However, since Ridge sacrifices its unbiasedness for the smaller regression coefficient variance (a.k.a a James-stein-type estimator), extra difficulty is brougt to the statistical tests and the post-estimation inference of Ridge. To reduce the grouping effect and obtain a stable variable-selection result, cross-validated group lasso and cross-validated elastic net (CV-en) are introduced \citet{zou2005regularization, friedman10}. By grouping the highly correlated variables together (i.e. they will be dropped out or included as a group), group lasso improves the robustness of lasso to the grouping effect. However, group lasso relies on manual grouping of variables, which heavily relies on the accuracy of the field knowledge. On the other hand, even though \citet{zou2005regularization} and \citet{jia2010model} show that in some cases CV-en improves the stability and accuracy of lasso variable selection, \citet{jia2010model} also show that the improvement is mariginal and ``when the lasso does not select the true model, it is more likely that the elastic net does not select the true model either.'' 

As the latest attempt to improve the accuracy and stability of variable-selection, \citet{ning2019solar} propose the \emph{subsample-ordered least-angle regression (solar)}, a ultrahigh dimensional redesign of lasso. Solar relies on the average $\mathcal{L}_0$ solution path computed across subsamples and largely alleviates several known high-dimensional issues with least-angle regression and $\mathcal{L}_1$ shrinkage.  Usinge examples based on directed acyclic graphs, we illustrate the advantages of  solar in comparison to least-angle regression, forward regression and variable screening. Simulations demonstrate that, with the same computation load, solar yields substantiali mprovements over two lasso solvers (least-angle regression for lasso and coordinate-descent) in terms of the sparsity (37-64\% reduction in the average number  of selected variables), stability and accuracy of variable selection. Simulations also demon-strate that solar enhances the robustness of variable selection to different settings of the irrepresentable condition and to variations in the dependence structures assumed in regression analysis.

\subsection{Main results}

In this paper, we compare the variable-selection performance of solar with lasso and CV-en in two real-world datasets, the prostate cancer data and Sydney house price data. The prostate cancer data is an small-size, industry-standard data for testing variable-selection and prediction performance of new estimators/algorithms in machine learning and biostatistics. In prostate cancer data, due to the heavy muticollinearity among all explanatory variables, lasso randomly drops an important variable and returns counterintuitive results; by contrast, solar and CV-en includes all important variables and return stable variable-selection results that are consistent with biostatistics theory. Alongside with simulations in last chapter, the performance of solar on prostate cancer data confirms the advantage of solar over lasso from the perspective of variable-selection stability and accuracy.

We also apply solar, lasso and CV-en to Sydney house price data. Compared with the prostate cancer data, Sydney house price data has more variables, more severe multicollinearity issue and more complicated dependence structure. As a a result, lasso and CV-en lose their sparsity on Sydney house price data, selecting respectively 44 and 57 variables out of 57 variables. By contrast, variables selected by solar still remains sparse (9 out of 57 variables) and intuitive. By dropping 48 variables, solar only reduces $R^2$ and $\overline{R}^2$ by a very small margin. By conducting the post-selection inference on the variable-selection result of solar, we further corrects the possible variable-selection issue in solar caused by multicollinearity and grouping effect.


\section{Solar variable selection on the real-world data with small $n$ and $p$}

In last chapter we introduce the solar algorithm, which is specifically designed for high-dimensional variable-selection with severe multicollinearity. In the simulation section of last chapter, it has been illustrated that (i) solar outperforms lasso from the perspective of variable-selection stability and accuracy; (ii) solar can still successfully identify informative variable with harsh IRC settings. In this section, we verifies the advantage of solar on the prostate-cancer data \citep{friedman2001elements}, a representative data with small $p$, small $n$ and severe multicollinearity and grouping effect.

\subsection{Data description}

The prostate-cancer data is collected for the prediction of the prostate cancer aggression. As shown in table~\ref{table:cancer_variable}, in this data we have $9$ medical test scores and $97$ prostate-cancer patients with positive and conclusive diagnoses. Among all variables, `lspa' is the log PSA test score, which for decades the U.S. Food and Drug Administration (FDA) have been using to measure the cancer aggression of a prostate cancer patient. For a prostate cancer patient, the higher the PSA score is, the more aggressive the cancer is. As a result, `lspa' is the response variable of the regression. The other variables in the data are used as covariates of the regression, which are different medical test results and also frequently used in the prostate-cancer diagnosis. Among all covariates in this data, `gleason' and `pgg45' are pathologically most relevant to the aggression of cancer. `gleason' is the current Gleason test score -- a score that ranges from 1 to 10 --  and is another standard FDA test score for the prostate cancer aggression. The higher the Gleason score is, the more aggressive the cancer is. Likewise, `pgg45' is the percentage of 4 or 5 Gleason socres that were recorded over the history (not including the current Gleason score). The Gleason score and the PSA score are major tools for prostate cancer diagnosis. From the perspective of the pathology theory, all covariates in the data are relevant to the prostate cancer diagnosis. By using these variables, we aim to predict the aggression of the prostate cancer as accurately as possible.
\begin{table}[H]
 \caption{Variables in prostate-cancer data \label{table:cancer_variable}}
 \begin{tabular}{l|l}
  \toprule
  name    & description \\
  \bottomrule
  lpsa    & the log PSA score                                            \\
  lcavol  & the log cancer volume                                        \\
  lweight & the log weight of prostate                                   \\
  age     & age of the patient                                           \\
  lbph    & the log amount of benign prostatic hyperplasia               \\
  svi     & the presence of seminal vesicle invasion (binary)            \\
  lcp     & the log amount of capsular penetration                       \\
  gleason & the current Gleason score (most cancers score 3 or higher)   \\
  pgg45   & the percentage of 4 or 5 Gleason scores that were recorded over the patient history \\
  \bottomrule
 \end{tabular}
\end{table}

However, due to the complicated dependence structure in this biostatistics data, the multicollinearity issue is severe among the covariates. As shown in table~\ref{table:corr_cancer}, \{lcavol, svi, lcp, gleason, pgg45\} are highly correlated with one another, implying that the classical variable selection method may be overtrigged and randomly drop some of them due to the `grouping effect'. Despite the instability of the variable selection result, the high muticollinearity may weaken the accuracy of regression coefficients estimation of the ordinary least square (OLS) or maximum likelihood estimation (MLE) on this sample. The standard error of each regression coefficiet estimate will be unnecessarily large, implying that the regression coefficiet estimate may change dramtically across samaples.

\begin{table}[H]
  \caption{Correlation table of all covariates in prostate cancer data \label{table:corr_cancer}}
  \begin{tabular}{l|lllllllll}
  \toprule
          & lcavol   & lweight  & age      & lbph      & svi      & lcp      & gleason  & pgg45 \\
  lcavol  & 1        &          &          &           &          &          &          &       \\
  lweight & 0.280521 & 1        &          &           &          &          &          &       \\
  age     & 0.225    & 0.347969 & 1        &           &          &          &          &       \\
  lbph    & 0.02735  & 0.442264 & 0.350186 & 1         &          &          &          &       \\
  svi     & 0.538845 & 0.155385 & 0.117658 & -0.085843 & 1        &          &          &       \\
  lcp     & 0.67531  & 0.164537 & 0.127668 & -0.006999 & 0.673111 & 1        &          &       \\
  gleason & 0.432417 & 0.056882 & 0.268892 & 0.07782   & 0.320412 & 0.51483  & 1        &       \\
  pgg45   & 0.433652 & 0.107354 & 0.276112 & 0.07846   & 0.457648 & 0.631528 & 0.751905 & 1      \\
  \bottomrule
  \end{tabular}
\end{table}

\subsection{Variable-selection results on prostate cancer data}

Table~\ref{table:variable_selected_cancer} shows the variable-selection comparison of solar, lasso solvers and CV-en. Since most of the covariates have been shown pathologically relevant to the severity of prostate cancer, all variable-selection methods select a similar combination of variables into the regression model. Most of the selected variables make perfect sense in pathology. For example, benign prostatic hyperplasia (variable `lbph') and inflammation may also cause PSA score to increase significantly. As a result, the inclusion of `lbph', `svi' and `lcp' is intuitive.

However, the only difference between lasso and solar/CV-en is the variable `gleason'. Two lasso solvers include `pgg45' instead of `gleason'. For prostate cancer patients with positive and conclusive diagnosis, this variable-selection result seems to suggest that the current PSA score -- an accurate measure of the current cancer aggression -- is not relevant to the current Gleason score, another accurate measure of the current cancer aggression. By contrast, it suggests that the current PSA score is relevant to `pgg45', the historical gleason values. This variable-selection result seems very counterintuitive. Consider a prostate cancer patient that just has never had any positive diagnosis before, his `pgg45' value will be $0$ while the corresponding values of `gleason' and `lpsa' would be high. In prostate cancer data, we have $35$ patients with $\mathrm{pgg45} = 0$ but $\mathrm{gleason} = 6$. Hence, for these patients `pgg45' is not useful for prostate cancer diagnosis.

 However, both CV-en and solar includes `gleason'. From the perspective of variable selection, CV-en is more likely to be robust to muticollinearity. As a result, the stability and accuracy of solar variable-selection is confirmed by the fact that CV-en and solar select the same variables. Nevertheless, we still need to investiage whether the drop-out of `gleason' in lasso is due to high multicollinearity and grouping effect.

\begin{table}[H]
 \centering
 \caption{Variables selected by solar, CV-lars-lasso and CV-cd in prostate cancer data. \label{table:variable_selected_cancer}}
 \begin{tabular}{ll.}
  \toprule
                & Variables selected                                   & \multicolumn{1}{c}{Total} \\
  \midrule
  solar         & lcavol, lweight, age, lbph, svi, lcp, gleason, pgg45 & 8   \\
  CV-en         & lcavol, lweight, age, lbph, svi, lcp, gleason, pgg45 & 8   \\
  CV-lars-lasso & lcavol, lweight, age, lbph, svi, lcp, pgg45          & 7   \\
  CV-cd         & lcavol, lweight, age, lbph, svi, lcp, pgg45          & 7   \\
  \bottomrule
 \end{tabular}
\end{table}

To investigate the grouping effect in the prostate cancer data, first we report the average $L_0$ solution path of solar in table~\ref{table:solar_cancer}. As shown in the table, on average, `gleason' and `pgg45' are included at the end of lars at each subsample, where `gleason' is included by lars before `pgg45'. As shown in last chapter, the $\widehat{q}_j$ value of a variable measure the stage that forward regression, on average, includes a variable. A varibale with a larger $\widehat{q}_j$ value, on average, will be included into forward regression at an earlier stage, implying that it is more likely to be an informative variable. As a result, solar suggests that on average `gleason' is more likely to be informative than `pgg45'. Compared to the variable-selection result of lasso, the result of solar makes more sense in pathology, especially for the 35 patients with $\mathrm{pgg} = 0$ in the data.

\begin{table}[H]
 \centering
 \caption{Variables in $Q(c) = \{ \mathbf{x}_j | \widehat{q}_j \geqslant c \}$ ($c^*$ in red) \label{table:solar_cancer}}
 \begin{tabular}{ll}
  \toprule
  \multicolumn{1}{c}{$c$} & Variables in $Q(c)$                             \\
  \midrule
  1                       & lcavol                                          \\
  0.844                   & lcavol, lweight                                 \\
  0.622                   & lcavol, lweight, age                            \\
  0.511                   & lcavol, lweight, age, lbph                      \\
  0.444                   & lcavol, lweight, age, lbph, svi                 \\
  0.266                   & lcavol, lweight, age, lbph, svi, lcp, gleason   \\
  \textcolor{red}{0.088}  & \textcolor{red}{lcavol, lweight, age, lbph, svi, lcp, gleason, pgg45} \\
  \bottomrule
 \end{tabular}
 \label{table:solar_select}
\end{table}

Table~\ref{table:solar_cancer} also shows that `pgg45' is included right after `gleason'. We know from last chapter that the $\widehat{q}_j$ value of a variable can be seen as the conditional relevance to $Y$. Hence, two highly correlated variables may have the similar relevance to $Y$, implying that they may be ranked close to each other in the average $L_0$ solution path. As a result, the location of `gleason' and `pgg45' in table~\ref{table:solar_cancer} suggests that, conditional on \{lcavol, lweight, age, lbph, svi, lcp\}, `gleason' and `pgg45' may be highly correlated to each other. To validate that hypothesis, first we report the marginal coorelation table of gleason to each other variable in the data (table~\ref{table:corr_gleason}). Table~\ref{table:corr_gleason} verifies the multicollinearity is severe for `gleason' in the prostate cancer data, which may potentially lead to the IRC violation. In this case, due to the sampling randomness and multicollinearity, lasso solvers may randomly drop gleason out of the regression even though `gleason' may be informative.

\begin{table}[H]
 \centering
 \caption{Marginal correlations to gleason \label{table:corr_gleason}}
 \begin{tabular}{llllllll}
  \toprule
          & lcavol & lweight & age & lbph  & svi & lcp & pgg45 \\
  \midrule
  $\mathrm{corr} \left( \; \cdot \; , \mathrm{gleason} \right)$ & 0.432  & 0.057   & 0.269 & 0.078 & 0.320 & 0.515 & 0.752 \\
  \bottomrule
 \end{tabular}
\end{table}

As shown in last chapter, IRC is vital in lasso regression. If IRC is violated, lasso variable selection may not be reliable, resulting in the inclusion of redundant variables and the exclusion of informative variables. Based on the pathological intuition and variable selection result of solar and CV-en, `gleason' is likely to be informative. Hence, to check if IRC is violated with respect to `gleason', we standardize all variables and estimate equation~(\ref{IRC_OLS:cancer}),
\begin{align}
 \mathrm{gleason} = \alpha_0 & + \alpha_1 \cdot \mathrm{lcavol} + \alpha_2 \cdot \mathrm{lweight} + \alpha_3 \cdot \mathrm{age} + \alpha_4 \cdot \mathrm{lbph} \notag \\
                             & + \alpha_5 \cdot \mathrm{svi} + \alpha_6 \cdot \mathrm{lcp} + \alpha_7 \cdot	\mathrm{pgg45} + e,
 \label{IRC_OLS:cancer}
\end{align}
and check the $L_1$ norm of its regression coefficient, which can be found in Table~\ref{table:IRC_OLS_cancer}.

\begin{table}[H]
  \centering
  \caption{OLS report of \ref{IRC_OLS:cancer} \label{table:IRC_OLS_cancer}}
  \begin{tabular}{rlrl}
    \toprule
    No. Observations: & 97      & AIC:                & 202.6    \\
    R-squared:        & 0.595   & BIC:                & 223.2    \\
    Adj. R-squared:   & 0.563   & F-statistic:        & 18.68    \\
    Dep. Variable:    & gleason & Prob (F-statistic): & 4.24e-15 \\
    \bottomrule
  \end{tabular}
  \begin{tabular}{r....}
                 & \mathrm{coef} & \mathrm{std \; err} & \mathrm{t}  & P>|t|   \\
    const        & 0             & 0.067               & 0           & 1.000   \\
    lcavol       & 0.1726        & 0.096               & 1.796       & 0.076   \\
    lweight      & -0.0916       & 0.081               & -1.134      & 0.260   \\
    age          & 0.0701        & 0.078               & 0.903       & 0.369   \\
    lbph         & 0.0261        & 0.079               & 0.329       & 0.743   \\
    svi          & -0.1080       & 0.094               & -1.154      & 0.252   \\
    lcp          & 0.0429        & 0.119               & 0.362       & 0.718   \\
    pgg45        & 0.6878        & 0.091               & 7.587       & 0.000   \\
    \bottomrule
  \end{tabular}
\end{table}

Table~\ref{table:IRC_OLS_cancer} reports the OLS result of \ref{IRC_OLS:cancer}. In this regression, the response variables and all covariates are standardized, implying that the regression coefficent actually represent the conditional correlation between a covariate and the response variable. Table~\ref{table:IRC_OLS_cancer} shows that around 60\% of the variation of gleason can be explained by \{lcavol, lweigh, age, lbph, svi, lcp, pgg45\}. Moreover, controlling the other covariates in this regression, the conditional correlation between `pgg45' and `gleason' is around $0.7$. This clearly shows that we have the grouping effect problem among \{lcavol, lweigh, age, lbph, svi, lcp, pgg45, gleason\}.

Even worse, table~\ref{table:IRC_OLS_cancer} also confirms the possible violation of IRC on `gleason'. As shown in the table, $\sum_{\forall i \neq 0}\left\vert \hat{\alpha}_i \right\vert$ in~(\ref{IRC_OLS:cancer}) is around 1.1. This implies that, even though `gleason' seems to be an informative variable in pathology and is included by solar and CV-en, lasso will still drop it out. Moreover, the dropout of `gleason' may be a random decision due to the sampling randomness and the grouping effect among \{lcavol, age, svi, lcp, pgg45, gleason\}. Put in another way, if we collect another sample from the same population and re-apply lasso on it, lasso is likely to randomly drop another variable from \{lcavol, age, svi, lcp, pgg45, gleason\}, probably `pgg45' due to high sample correlation. As a result, it is diffcult to interpret the variable-selection result of lasso.

The variable-selection comparison among lasso, CV-en and solar confirms the advantage of solar, which we have demonstrated in last chapter. The dependence structure in prostate cancer data may not be very complicated due to relatively small $p$ (9 variables in this data). In such scenario, lasso is affected by grouping effect and may be not reliable. By contrast, solar and CV-en are both robust to the grouping effect and return a intuitive and stable variable-selection result. In next section, to check the robustness of each variable selection method to grouping effect, we apply CV-en, solar and lasso on Sydney house price data, the one with larger $p$ and, hence, much more complicated dependence structures.


\section{Solar variable selection on Sydney house data}

In last section, we demonstrate the advantage of solar in prostate cancer data, which is an industry-representative of data with small $p$ and severe grouping effect. Due to the well-implemented laboratory experiments/tests and the accurate procedure of clinical data collection, many biostatistical datasets are known to be with clear and well-define dependence structures. Also, due to the limitation of the value of $p$, the dependence structure in prostate cancer data is still managable. In this section, we focus on the performance of solar, lasso and CV-en on Sydney house price data, a data with larger $p$, much more complicated dependence struture and more severe grouping effect.

\subsection{Data description}

Sydney house price data is collected for the price prediction of all the second-hand houses on the market of Mid and East Sydney, 2010. As shown in Table~\ref{table:house_variable} (at the end of the paper), in Sydney house price data we have 57 covariates, which can be classified into 3 categories: the features of the house, distance to key locations (public transport, shopping, etc), local school quality and the localized socio-economic data. The features of the house is reported in the real-estate transactions alongside with the house price. The distance of each house to the nearest key locations are computed in QGIS -- a open-source geographical information system -- based on the GPS location of each house and the geo-data collected from Department of Land and Natural Resources, New South Wales, Australia. The ICSEA score -- an indication of the socio-educational backgrounds of students-- is collected from Australian Curriculum, Assessment and Reporting Authority (ACARA). The variables about local school quality (the average examination scores) is collected from Department of Education, New South Wales, Australia.

To check the possible multicollinearity and grouping effect in Sydney house price data, we also need to check the pairwise correlation table among all covariates. Due to the size of the table, we report it as an additional CSV file, which shows that multiple covariates in the Sydney houes price data are highly corelated with one another. As we expected, the possible grouping effect and multicollinearity issue in Sydney house price data may be much worse that the prostate cancer data, which implies a much more complicated dependence structure.

\subsection{Variable selection results}

Table~\ref{table:variable_selected_house} shows the selection results of solar, lasso and CV-en. In data with complicated dependence structures and severe multicollinearity, both lasso and CV-en lose sparsity of variable selection. Two lasso solvers only manage to drop 7 out of 55 variables and CV-en select all 57 variables. This is consistent with the finding of \citet{jia2010model}. ``When the lasso does not select the true model, it is more likely that the elastic net does not select the true model either.'' Heurestically increasing the value of $\lambda$ in lasso (e.g., one-se rule) may potentially improve the sparisty of lasso. However, this may leads to lasso randomly dropping variables due to the grouping effect. On the other hand, CV-en is designed to tolerate multicollinearity and grouping effect, returning a sparse and stable regression model. However, due to the complicated dependence structure, CV-en completely fails to accomplish variable selection. Conclusively, $L_1$ shrinkage methods fail to simultaneously maintain sparsity and stability in this data. By contrast, as a variable-selection algorithm robust to the complication of the dependence structure, solar still returns a very sparse regression model, only $9$ variables selected out of $57$.

\begin{table}[H]
 \centering
 \caption{Variables selected by solar, CV-lars-lasso and CV-cd in house price data.\label{table:variable_selected_house}}
 \begin{tabular}{ll.}
  \toprule
          & Variables selected & \multicolumn{1}{c}{Total} \\
  \midrule
  solar                    & Mean\_mortgage\_repay\_monthly, Mean\_rent\_weekly, Mean\_Tot\_fam\_inc\_weekly, & 9                 \\
  \phantom{CV-lars-lasso/} & Bedrooms, Baths, Parking, Beach, Gaol, ICSEA  &  \\
  CV-lars-lasso/           & Lang\_spoken\_home\_Eng\_only\_P, Australian\_citizen\_P,  Mean\_age\_persons, & 44      \\
  CV-cd                    & Mean\_mortgage\_repay\_monthly, Mean\_rent\_weekly,
  Mean\_Tot\_fam\_inc\_weekly,  \\
  \phantom{CV-lars-lasso/} & Average\_num\_psns\_per\_bedroom, Average\_household\_size, TVO2009, Suburb\_Area,          \\
  \phantom{CV-lars-lasso/} & AreaSize, Bedrooms, Baths, Parking, Airport, Beach, Cemetery, ChildCare, Club,                \\
  \phantom{CV-lars-lasso/} & GolfCourt, High, Lib, Museum, Park, Police, PreSchool, PrimaryHigh, Primary,           \\
  \phantom{CV-lars-lasso/} & RailStat, Rubbish, SportsCenter, SportsCourtField, Swimming, Tertiary, DistBound,            \\
  \phantom{CV-lars-lasso/} & ICSEA, ReadingY3, WritingY3, SpellingY3, GrammarY3, NumeracyY3, WritingY5,                      \\
  \phantom{CV-lars-lasso/} & SpellingY5, GrammarY5    \\
  CV-en                    & Tot\_P\_P, Lang\_spoken\_home\_Eng\_only\_P, Australian\_citizen\_P, Mean\_age\_persons,& 57                           \\
  \phantom{CV-lars-lasso/} & Mean\_mortgage\_repay\_monthly, Mean\_Tot\_prsnl\_inc\_weekly, Mean\_rent\_weekly,                      \\
  \phantom{CV-lars-lasso/} & Mean\_Tot\_fam\_inc\_weekly, Average\_num\_psns\_per\_bedroom, Average\_household\_size,                 \\
  \phantom{CV-lars-lasso/} & TVO2010, TPO2010, TVO2009, TPO2009, Suburb\_Area, AreaSize, Bedrooms,                                                                   \\
  \phantom{CV-lars-lasso/} & Baths, Parking, Airport, Beach, Cemetery, ChildCare, CommunityFacility, Club,                                                    \\
  \phantom{CV-lars-lasso/} & Gaol, GeneralHospital, GolfCourt, High, Lib, MedCenter, Museum, Park, PO, Police,                                                           \\
  \phantom{CV-lars-lasso/} & PreSchool, PrimaryHigh, Primary, RailStat, Rubbish, Sewage, SportsCenter,                                                               \\
  \phantom{CV-lars-lasso/} & SportsCourtField, Swimming, Tertiary, DistBound, ICSEA, ReadingY3, WritingY3,                                                       \\
  \phantom{CV-lars-lasso/} & SpellingY3, GrammarY3, NumeracyY3, ReadingY5, WritingY5, SpellingY5,                                                                 \\
  \phantom{CV-lars-lasso/} & GrammarY5, NumeracyY5                            \\
  \bottomrule
 \end{tabular}
\end{table}

Table~\ref{table:reg_coef_house} reports the regression coefficents of OLS, lasso solvers and solar. Due to the dimensionality, we only focus on the regression coefficents of variables selected by solar. Since the lasso regression coefficiets are under $L_1$ penalty, they are biased toward $0$ and typically smaller than the OLS regression coefficents, implying that the magnitude of lasso regression coefficents are not particularly helpful for the mariginal effect evaluation. Since (i) the regression coefficents of the elastic net are under the composite $L_1$-$L_2$ penalty; (ii) the elastic net penalty (and hence the bias in elastic net regression coeficients) is more complicated than lasso, we skip the regression coefficent value of elastic net in table~\ref{table:reg_coef_house}. Even though solar and OLS regression coefficents are both unbiased under regularity conditions, in this scenario the solar regression coefficients are still preferred due to the sparsity of the solar regression model, which reduces the severity of the curse of dimensionality and only returns the most important variables in regression modelling.

\begin{table}[H]
  \centering
  \caption{Regression coefficiet estimates of OLS, CV-lars-lasso, CV-cd and solar.\label{table:reg_coef_house}}
  \scalebox{0.75}{%
    \begin{tabular}{l....}
      \toprule
                                       & \textrm{OLS}  & \textrm{CV-lars-lasso} & \textrm{CV-cd} & \textrm{solar}  \\
      \midrule
      Mean\_mortgage\_repay\_monthly & 133.67        & 141.32        & 141.10        & 185.99       \\
      Mean\_rent\_weekly             & 264.35        & 285.14        & 284.55        & 370.76       \\
      Mean\_Tot\_fam\_inc\_weekly    & 59.04         & 58.97         & 59.30         & 66.57        \\
      Bedrooms                         & 165,639.00    & 166,335.90    & 166,330.74    & 169,510.52   \\
      Baths                            & 210,101.80    & 210,667.43    & 210,667.56    & 209,626.52   \\
      Parking                          & 97,790.57     & 95,993.28     & 96,022.85     & 97,623.23    \\
      Beach                            & -5,029,682.00 & -3,546,045.17 & -3,560,356.02 & -796,281.77   \\
      Gaol                             & 1,614,215.00  & 0.00          & 0.00          & 1,909,369.80 \\
      ICSEA                            & 838.54        & 813.08        & 814.66        & 1,756.92     \\
      \bottomrule
    \end{tabular}}
\end{table}

As shown in the simulation of last chapter and the prostate cancer data, the accuracy and stability of variable selection may be reduced by multicollinearity and grouping effect embedded in the data, especially when the potential dependence structure is large. To investigate if solar variable selection is affected by grouping effect, we first report the average $L_0$ solution path in house price data, shown in Table~\ref{table:solar_select}.

\begin{table}[H]
  \centering
  \caption{Variables in $Q(c) = \{ \mathbf{x}_j | \widehat{q}_j \geqslant c \}$ ($c^*$ in red)}
  \begin{tabular}{ll}
    \toprule
    \multicolumn{1}{c}{$c$} & Variables in $Q(c)$ \\
    \midrule
    1     & Baths                                                       \\
    0.977 & Baths, Mean\_mortgage\_repay\_monthly                     \\
    0.955 & Baths, Mean\_mortgage\_repay\_monthly, Bedrooms           \\
    0.956 & Baths, Mean\_mortgage\_repay\_monthly, Bedrooms, Mean\_Tot\_fam\_inc\_weekly \\
    0.933 & Baths, Mean\_mortgage\_repay\_monthly, Bedrooms, Mean\_Tot\_fam\_inc\_weekly, ICSEA \\
    0.911 & Baths, Mean\_mortgage\_repay\_monthly, Bedrooms, Mean\_Tot\_fam\_inc\_weekly, ICSEA,\\
    \phantom{0.911} & Mean\_rent\_weekly \\
    0.888 & Baths, Mean\_mortgage\_repay\_monthly, Bedrooms, Mean\_Tot\_fam\_inc\_weekly, ICSEA,\\
    \phantom{0.911} & Mean\_rent\_weekly, Parking \\
    \textcolor{red}{0.866}  & \textcolor{red}{ Baths, Mean\_mortgage\_repay\_monthly, Bedrooms, Mean\_Tot\_fam\_inc\_weekly, ICSEA,} \\
    \phantom{0.911} & \textcolor{red}{ Mean\_rent\_weekly, Parking, Gaol, Beach } \\
    0.844 & Baths, Mean\_mortgage\_repay\_monthly, Bedrooms, Mean\_Tot\_fam\_inc\_weekly, ICSEA, \\
    \phantom{0.911} & Mean\_rent\_weekly, Parking, Gaol, Beach, ChildCare \\
    \bottomrule
  \end{tabular}
  \label{table:solar_select}
\end{table}

Table~\ref{table:solar_select} shows that, variable `Gaol' and `Beach' have similar $\widehat{q}_j$ values (around 0.866), implying they may be high corrleated. Also, `ChildCare' is included right after `Beach' and `Gaol' with $\widehat{q}_j = 0.844$, suggesting that `Gaol' may also be correlated with `ChildCare' and `Gaol'. Hence, we are going to check the group of variables that is highly correlated to `Gaol', which is reported in table~\ref{table:corr_gaol}. Table~\ref{table:corr_gaol} shows that, `Airport', `Rubbish' and `ChildCare' are all highly correlated to `Gaol' (pairwise correlation larger than 0.5). As a result, such high correlation may trigger the grouping effect of variable selection and potentially violate IRC.

\begin{table}[H]
 \centering
 \caption{Marginal correlations to Gaol (absolute value larger than 0.5) \label{table:corr_gaol}}
 \begin{tabular}{lllll.}
  \toprule
  & ChildCare & Airport & Rubbish & Beach \\
  \midrule
  $\mathrm{corr} \left( \; \cdot \; , \mathrm{Gaol} \right)$ & 0.756     & 0.715   & 0.671   & 0.528 \\
  \bottomrule
 \end{tabular}
\end{table}

Based on table~\ref{table:corr_gaol}, it is necessary to check if the IRS with respect to `Gaol' is violated. As a result, we standardize all variables and esimate regression equation~(\ref{OLS:Gaol_others}),
\begin{equation}
 \mathrm{Gaol} = \gamma_0 + \gamma_1 \cdot \mathrm{Airport} + \gamma_2 \cdot \mathrm{ChildCare} + \gamma_3 \cdot \mathrm{Rubbish} + \gamma_4 \cdot \mathrm{Beach} + e.
 \label{OLS:Gaol_others}
\end{equation}
The OLS result of~(\ref{OLS:Gaol_others}) is reported in table~\ref{table:corr_gaol}.

\begin{table}[H]
  \centering
  \caption{OLS report of \ref{OLS:Gaol_others} \label{table:corr_gaol}}
  \begin{tabular}{r.r.}
    \toprule
    No. Observations: & 11974 & F-statistic:       & 23110  \\
    R-squared:        & 0.885 & Prob(F-statistic): & 0      \\
    Adj. R-squared:   & 0.885 & Df Model:          & 4      \\
    \bottomrule
  \end{tabular}
  \begin{tabular}{r....}
                    & coef   & std err & t       & P>|t| \\
    const           & 0      & 0.003   & 0       & 1.000 \\
    Airport         & 0.4488 & 0.011   & 41.063  & 0.000 \\
    ChildCare       & 0.3276 & 0.006   & 56.908  & 0.000 \\
    Rubbish         & 0.0373 & 0.010   & 3.849   & 0.000 \\
    Beach           & 0.5522 & 0.003   & 174.257 & 0.000 \\
    \bottomrule
  \end{tabular}
\end{table}

As table~\ref{table:corr_gaol} shows, the collinearity between Gaol and \{ChildCare, Airport, Rubbish, Beach\} is very severe. Almost 90\% of the variation of Gaol can be explained by \{ChildCare, Airport, Rubbish, Beach\} and $\sum_{\forall i \neq 0} \gamma_i = 1.35$ in (\ref{OLS:Gaol_others}). Compared with what we have in prostate cancer data, the grouping effect and multicollinearity in Sydney house price data is much more severe. As a result, even solar may not be completely immune of the severe grouping effect. This implies that, even if one of the variables in \{ChildCare, Airport, Rubbish, Beach, Gaol\} may be informative, it is very likely that variable-selection algorithms fail to identify the informative variable in that group. As a result, the inclusion of `Gaol', `ChildCare' and `Beach' may actually serve as a placeholder of the group \{ChildCare, Airport, Rubbish, Beach, Gaol\} in the variable-selection result. To avoid the misleading of the grouping effect, it is statistically reasonable to replace \{ChildCare, Beach, Gaol\} in the variable-selection result of solar with \{ChildCare, Airport, Rubbish, Beach, Gaol\}. We referred to the revised variable-selection result of solar as the `rectified solar selection'.

There is an empirical reason why \{Gaol, ChildCare, Airport, Rubbish, Beach\} are highly correlated with each other. All observations in the house price data is collected in East and Mid Sydney, Australia at 2010. As shown in Goolge map, in East and Mid Sydney, the gaol (Long Bay correctional complex), childcare center (e.g., Blue Gum Cottage Child Care, Alouette Child Care), airport (Kingsford-Smith Airport) and rubbish incenerators (e.g., Malabar Wastewater Treatment Plant, Sydney Desalination Plant, Cronulla Wastewater Treatment Plant, Bondi Wastewater Treatment Plant) all concentrate at the southeast coastline of East Sydney, which explains the collinearity of those variables.

At the end of this subsection, we estimate OLS regressions based on the variable-selection results of lasso, CV-en and solar and compare their $R^2$ and $\overline{R}^2$. For completeness, we also estimate an OLS regression based on the `rectified solar selection' (equation~(\ref{OLS:postsolar})) and compare its performance with solar regression results (equation~(\ref{OLS:solar})).
\begin{align}
 \mathrm{Price} = \beta_0 & + \beta_1 \cdot \mathrm{Mean\_mortgage\_repay\_monthly} + \beta_2 \cdot \mathrm{Mean\_rent\_weekly} \notag                                \\
                          & + \beta_3 \cdot \mathrm{Mean\_Tot\_fam\_inc\_weekly} + \beta_4 \cdot \mathrm{Bedrooms} + \beta_5 \cdot \mathrm{Baths} \label{OLS:solar}     \\
                          & + \beta_6 \cdot \mathrm{Parking} + \beta_7 \cdot \mathrm{Beach} + \beta_8 \cdot \mathrm{Gaol} + \beta_9 \cdot \mathrm{ICSEA} + u \notag       \\
 \mathrm{Price} = \beta_0 & + \beta_1 \cdot \mathrm{Mean\_mortgage\_repay\_monthly} + \beta_2 \cdot \mathrm{Mean\_rent\_weekly} \notag                                \\
                          & + \beta_3 \cdot \mathrm{Mean\_Tot\_fam\_inc\_weekly} + \beta_4 \cdot \mathrm{Bedrooms} + \beta_5 \cdot \mathrm{Baths} \label{OLS:postsolar} \\
                          & + \beta_6 \cdot \mathrm{Parking}  + \beta_7 \cdot \mathrm{Beach} + \beta_8 \cdot \mathrm{Airport} + \beta_9 \cdot \mathrm{ChildCare} \notag   \\
                          & + \beta_{10} \cdot \mathrm{Rubbish} + \beta_{11} \cdot \mathrm{ICSEA} + u \notag
\end{align}
The comparison results are summarized in table~\ref{table:house_compare}. In this case, lasso clearly does a better job on variable selection than CV-en. With slightly better sparsity, the variables selected by lasso produce the same $\overline{R}^2$ with respect to price as does CV-en. Note that the variables that lasso drops are highly correlated with those that lasso selects, which implies another potential problem of the grouping effect. However, the results of CV-en and lasso may be considered not sparse enough for many economics and ecnonometrics analysis. By contrast, the rectified solar selection delete 46 variables and returns a very sparse model. Moreover, the sparisty is accomplished by reducing $\overline{R}^2$ by only 0.03. As a result, it clearly shows that, rectified solar selection balances $\overline{R}^2$ and number of variables better than lasso and CV-en in Sydney house price data.

\begin{table}[H]
 \caption{Post selection OLS of solar, CV-cd, CV-lars-lasso and CV-en \label{table:house_compare}}
 \begin{tabular}{llll}
  \toprule
                                    & number of variabels & $R^2$ & $\overline{R}^2$ \\
  \midrule
  solar (\ref{OLS:solar})           & 9                   & 0.494 & 0.493            \\
  rectified-solar (\ref{OLS:postsolar}) & 11                  & 0.514 & 0.514            \\
  CV-cd/CV-lars                     & 44                  & 0.548 & 0.546            \\
  CV-en                             & 57                  & 0.548 & 0.546            \\
  \bottomrule
 \end{tabular}
\end{table}

\section{Conclusion}

In this paper we demonstrate the performance of solar variable selection on different empirical data with severe multicollinearity issue and, hence, a severe grouping effect. As the competitor of solar, lasso is affected by the grouping effect and returns the unreliable variable-selection result; even though more robust to the grouping effect than lasso, CV-en loses the sparsity of the variable-selection result when $p$ gets large. By contrast, solar returns a stable and sparse variale-selection result and illustrates a better robustness to the grouping effect. As a result, the advantage of solar that we demonstrate in the simulation of last chapter is verified in empirical data.


\bibliographystyle{elsarticle-harv}
\bibliography{CVrefs}


\appendix

\begin{landscape}
  \begin{table}
    \caption{Variables in Sydney house price data \label{table:house_variable}}
    \begin{tabular}{ll|ll.}
      \toprule
      Airport             & Distance to the nearest airport              & Mean\_mortgage\_repay\_monthly & the meidan mortgage repayment  in the local SA1           &  \\
      Beach               & Distance to the nearest beach                & Mean\_Tot\_prsnl\_inc\_weekly  & the meidan personal income  in the local SA1              &  \\
      Cemetery            & Distance to the nearest cemetery             & Mean\_rent\_weekly             & the meidan rent in the local SA1                          &  \\
      ChildCare           & Distance to the nearest childcare center     & Mean\_Tot\_fam\_inc\_weekly    & the meidan family income in the local SA1                 &  \\
      CommunityFacility   & Distance to the nearest community facility   & Average\_num\_psns\_per\_bedroom & the aveage population to bedroom ratio in the local SA1 &  \\
      Club                & Distance to the nearest club                 & Average\_household\_size         & the aveage house size in the local SA1                    &  \\
      Gaol             & Distance to the nearest gaol                 & TVO2010      &  Total Violent Offences of the suburb at 2010  &  \\
      GeneralHospital     & Distance to the nearest general hospital     & TPO2010              & Total Property Offences of the suburb at 2010                    &  \\
      GolfCourt           & Distance to the nearest golf court           & TVO2009   &   Total Violent Offences of the suburb at 2009                             &  \\
      High                & Distance to the nearest high school          & TPO2009              & Total Property Offences of the suburb at 2009   &  \\
      Lib                 & Distance to the nearest library              & Suburb\_Area                     & Area of the local suburb                                  &  \\
      MedCenter           & Distance to the nearest medical center       & AreaSize                         & Area size of the house                                    &  \\
      Museum              & Distance to the nearest museum               & Bedrooms                         & Number of bedrooms in the house                           &  \\
      Park                & Distance to the nearest park                 & Baths                            & Number of bathrooms in the house                          &  \\
      PO                  & Distance to the nearest post office          & Parking                          & Number of parking space of the house                      &  \\
      Police              & Distance to the nearest police station       & ICSEA                            & the average ICSEA score for the local school catchment    &  \\
      PreSchool        & Distance to the nearest preschool            & ReadingY3                        & the average reading score of year 3                       &  \\
      PrimaryHigh         & Distance to the nearest PrimaryHigh          & WritingY3                        & the average writing score of year 3                       &  \\
      Primary             & Distance to the nearest primary school       & SpellingY3                    & the average spelling score of year 3                      &  \\
      RailStat            & Distance to the nearest rail station         & GrammarY3                        & the average grammar score of year 3                       &  \\
      Rubbish             & Distance to the nearest rubbish incenerator  & NumeracyY3                       & the average numeracy score of year 3                      &  \\
      Sewage              & Distance to the nearest sewage               & ReadingY5                        & the average reading score of year 5                       &  \\
      SportsCenter        & Distance to the nearest sports center        & WritingY5                        & the average writing score of year 5                       &  \\
      SportsCourtField & Distance to the nearest sports court/field   & SpellingY5                       & the average spelling score of year 5                      &  \\
      Swimming            & Distance to the nearest swimming court/filed & GrammarY5                        & the average grammar score of year 5                       &  \\
      Tertiary            & Distance to the nearest tertiary school      & NumeracyY5                       & the average numeracy score of year 5                      &  \\
      DistBound           & Distance to the nearest suburb Bound         &                                  &                                                           & \\
      \bottomrule
    \end{tabular}
  \end{table}
\end{landscape}

\end{document}